\documentclass[10pt,twocolumn,letterpaper]{article}

\usepackage{cvpr}              % To produce the CAMERA-READY version
\usepackage[dvipsnames]{xcolor}

\definecolor{cvprblue}{rgb}{0.21,0.49,0.74}
\usepackage[pagebackref,breaklinks,colorlinks,citecolor=cvprblue]{hyperref}

\usepackage{capt-of,xcolor,xspace,enumitem}
\usepackage{amsmath,amssymb,amsbsy,amsfonts,dsfont,pifont,bm,bbm,mathrsfs,mathtools,nicefrac}
\usepackage{algorithm,algpseudocode,listings}
\usepackage{booktabs,multirow,adjustbox,diagbox,threeparttable,tabularray}

\usepackage{orcidlink}
\usepackage{wrapfig}
\usepackage[capitalize]{cleveref}  % Should be loaded after 'hyperref', and works perfectly with 'subfigure'.
\crefname{section}{Sec.}{Secs.}
\Crefname{section}{Section}{Sections}
\crefname{table}{Tab.}{Tabs.}
\Crefname{table}{Table}{Tables}
\crefname{figure}{Fig.}{Figs.}
\Crefname{figure}{Figure}{Figures}
\crefname{equation}{Eq.}{Eqs.}
\Crefname{equation}{Equation}{Equations}
\DeclareMathOperator*{\argmin}{arg\,min}
\newcommand{\method}{\textcolor{black}{{\mbox{\texttt{AGAP}}}}\xspace}

\def\paperID{105} % *** Enter the Paper ID here
\def\confName{3DV\xspace}
\def\confYear{2025\xspace}

\title{Learning Naturally Aggregated Appearance for Efficient 3D Editing}

\author{
    Ka Leong Cheng\textsuperscript{1,2}, ~~
    Qiuyu Wang\textsuperscript{2}, ~~
    Zifan Shi\textsuperscript{1,2}, ~~
    Kecheng Zheng\textsuperscript{2}, ~~
    Yinghao Xu\textsuperscript{2,3}, ~~ \\[2pt] 
    Hao Ouyang\textsuperscript{2}, ~~ 
    Qifeng Chen\textsuperscript{1}\footnotemark[2]~, ~~ 
    Yujun Shen\textsuperscript{2}\footnotemark[2]
    \\[6pt]
    $^1$HKUST ~
    $^2$Ant Group ~
    $^3$Stanford
}

\begin{document}

\twocolumn[{
\renewcommand\twocolumn[1][]{#1}
\maketitle
\begin{center}
    \vspace{-15pt}
    \includegraphics[width=1.0\linewidth]{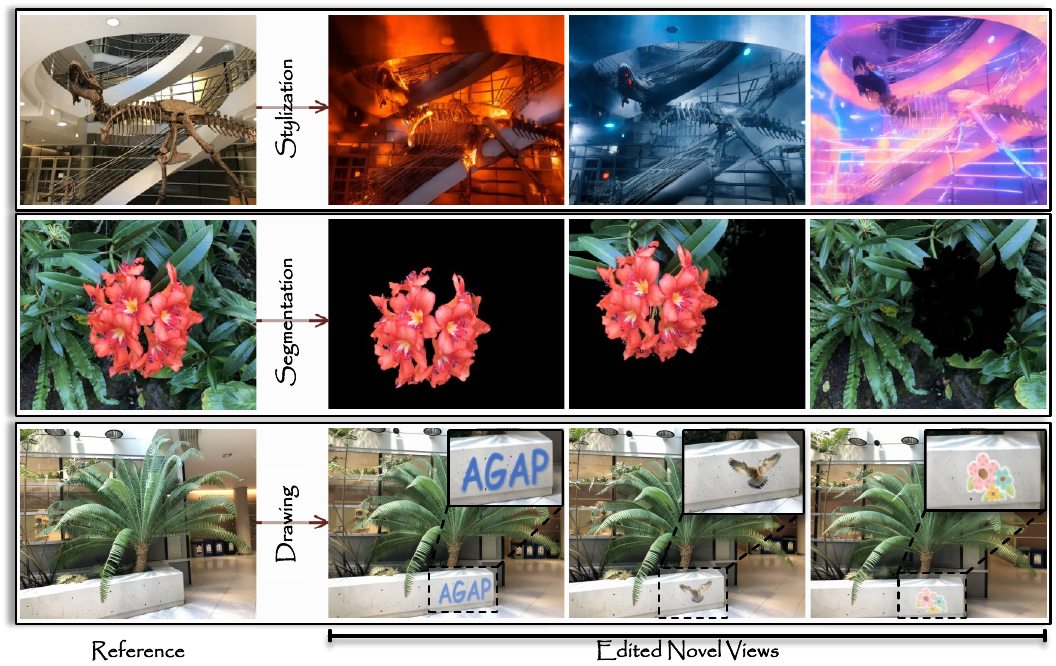}
    \vspace{-10pt}
    \captionsetup{type=figure}
    \caption{%         
        \method aggregates 3D appearance as natural 2D canonical images. With image processing tools, \method enables various ways of 3D editing without re-optimization, including (a) scene stylization, (b) instance segmentation, and (c) interactive drawing.
    }
    \label{fig:teaser}
\end{center}
}]

\begin{abstract}

Neural radiance fields, which represent a 3D scene as a color field and a density field, have demonstrated great progress in novel view synthesis yet are unfavorable for editing due to the implicitness.
This work studies the task of efficient 3D editing, where we focus on \textbf{editing speed} and \textbf{user interactivity}.
To this end, we propose to learn the color field as an explicit 2D appearance aggregation, also called canonical image, with which users can easily customize their 3D editing via 2D image processing.
We complement the canonical image with a projection field that maps 3D points onto 2D pixels for texture query.
This field is initialized with a pseudo canonical camera model and optimized with offset regularity to ensure the \textbf{naturalness} of the canonical image.
Extensive experiments on different datasets suggest that our representation, dubbed \method, well supports various ways of 3D editing (\textit{e.g.,} stylization, instance segmentation, and interactive drawing).
Our approach demonstrates remarkable efficiency by being at least 20$\times$ faster per edit compared to existing NeRF-based editing methods.
Project page is available at~\url{https://felixcheng97.github.io/AGAP/}.

\end{abstract}

\begin{table*}[t]
    % \vspace{-10pt}
    \small
    \caption{\method supports various editing cases. Our editing time per edit is significantly shorter thanks to the optimization-free editing pipeline. Note that the times listed below do not include the comparable pre-training and rendering times across methods.}
    \vspace{-8pt}
    \label{tab:functionality}
    \centering
    \SetTblrInner{rowsep=1.2pt}      % Row space.
    \SetTblrInner{colsep=10.0pt}      % Col space.
    \begin{tblr}{
        cells={halign=c,valign=m},   % Text alignment for all cells.
        column{1}={halign=l},        % Text alignment for the first column.
        hline{1,2,8}={1-6}{},        % Horizontal lines.
        hline{1,8}={1.0pt},          % Horizontal line width.
        vline{2,6}={1-7}{},          % Vertical lines.
        row{7}={bg=lightgray!50},
    }
    \                                            & Global stylization & Local stylization                & Segmentation                     & Drawing   & Editing time\footnotemark[3] \\
    ARF~\cite{zhang2022arf}                      & \ding{51}          & \ding{55}                        & \ding{55}                        & \ding{55} & 371s         \\
    Ref-NPR~\cite{zhang2023refnpr}               & \ding{51}          & \ding{55}                        & \ding{55}                        & \ding{55} & 514s         \\
    DFFs~\cite{kobayashi2022decomposing}         & \ding{55}          & \ding{51}                        & \ding{51}                        & \ding{55} & 516s         \\
    IN2N~\cite{haque2023instruct}                & \ding{51}          & $\boldsymbol{-}$\footnotemark[1] & \ding{55}                        & \ding{55} & $\sim$10000s \\
    GaussianEditor~\cite{chen2024gaussianeditor} & \ding{51}          & \ding{51}                        & $\boldsymbol{-}$\footnotemark[2] & \ding{55} & 1320s        \\
     Ours                                        & \ding{51}          & \ding{51}                        & \ding{51}                        & \ding{51} & 20s          \\
    \end{tblr}
    \vspace{-10pt}
\end{table*}

\footnotetext[1]{Possible but highly depends on prompts}
\footnotetext[2]{Not inherently support in the official implementation}
\footnotetext[3]{Editing time for stylization evaluated on a single A6000 GPU}

\section{Introduction}\label{sec:intro}

While recent advancements in 3D representations like neural radiance fields (NeRF)~\cite{mildenhall2020nerf} have shown impressive reconstruction capabilities for real-world scenes, the need for further progress in 3D editing arises as the desire to recreate and manipulate these scenes. The field of 3D editing has witnessed significant development in recent years. Traditional 3D modeling approaches~\cite{hoppe2006poisson,schonberger2016pixelwise,schnberger2016structure,xiang2021neutex} typically rely on reconstructing scenes using meshes. By combining meshes with texture maps, we can enable appearance editing during the rendering process. However, these methods usually face difficulties in obtaining detailed and regular texture maps, typically in complex scenes, hindering effective editing and user-friendliness.

Recent neural radiance fields offer high-quality scene reconstructions, but manipulating the implicit 3D representation embedded within neural networks is inherently non-straightforward. Existing NeRF-based editing approaches can be mainly divided into two categories: some methods like~\cite{chen2023neuraleditor,deng2021deformed,wang2021neus,yang2022neumesh,yuan2022nerfediting,zhang2023papr} target geometry editing, usually taking advantage of meshes, while the other~\cite{haque2023instruct,pang2023locally,kobayashi2022decomposing,zhang2022arf,zhang2023refnpr} focuses on 3D stylization using images or texts as guidance. However, re-optimizing the original NeRF models is necessary to incorporate the desired editing effects into the underlying 3D representation, resulting in time-consuming processes. Consequently, it is crucial to develop a user-friendly framework that can efficiently and effectively support various edits within a single model.

This paper introduces a novel editing-friendly representation \textbf{\method} with naturally \textbf{Ag}gregated \textbf{Ap}pearance for efficient 3D editing, consisting of a 3D density grid for geometry estimation and a canonical image plus a projection field for appearance modeling. Our method attempts to link the 3D representation with a natural 2D canonical representation. Concretely, a learnable canonical image is designed as the interface for editing, which aggregates the appearance by projecting the 3D radiance to a natural-looking image by the associated projection field. To ensure the naturalness of the aggregated canonical image with strong representation capacity, the projection field is carefully initialized using a pseudo canonical camera model and complemented by a learned view-dependent offset. The underlying 3D structure is modeled by an explicit 3D density grid.

\method supports various ways edits of a 3D scene in a user-friendly and efficient manner by applying different 2D image processing tools on the canonical images without re-optimization.~\cref{tab:functionality} summarizes the comparison of our method with existing methods in terms of editing functionalities and per-edit efficiency. We evaluate the effectiveness on various datasets, including LLFF~\cite{mildenhall2019llff}, DTU~\cite{jensen2014dtu}, NeUVF~\cite{ma2022nep}, Replica~\cite{straub2019replica,habtegebrial2022somsi}, IN2N~\cite{haque2023instruct}, NeRF-Synthetic~\cite{mildenhall2020nerf}, and Mip-NeRF 360~\cite{barron2022mipnerf360} datasets in various editing tasks which are scene stylization, instance segmentation, and texture editing (\textit{i.e.}, drawing). Experimental results show \method support various 3D editing tasks with on-par performance but at least 20$\times$ faster per edit.

\section{Related Work}\label{sec:related}

\noindent{\textbf{Implicit 3D representation.}}
3D modeling~\cite{achlioptas2018learning,ji2017surfacenet,kanazawa2018learning,sitzmann2018deepvoxels,wang2019mvpnet,wang2018pixel2mesh,liu2016learning,huang2018deepmvs} is pivotal in computer graphics and computer vision. Traditionally, explicit representations such as voxels and meshes have been employed for 3D shape modeling, but they often face challenges related to detail preservation and limited flexibility in processing. In contrast, implicit 3D representation like NeRF~\cite{mildenhall2020nerf}, SDF~\cite{park2019deepsdf,wang2021neus,yariv2021volume}, Occupancy Networks~\cite{mescheder2019occnet}, describing 3D scenes through continuous implicit functions, excels in capturing detailed geometry with improved fidelity. Many further works aim at improving NeRF in terms of various aspects, such as modeling capacity~\cite{barron2022mipnerf360,barron2021mipnerf}, generative modeling~\cite{wang2023benchmarking,schwarz2020graf,chan2022eg3d,gu2022stylenerf}, and camera pose estimation~\cite{lin2021inerf, wang2021nerf, lin2021bundle}. In particular, methods like DVGO~\cite{sun2022dvgo}, Plenoxels~\cite{fridovichkeil2022plenoxels}, InstantNGP~\cite{muller2022instantngp}, TensoRF~\cite{chen2022tensorf} focus on improving the convergence speed of volume rendering for 3D scenes by modeling the geometry and appearance with explicit grid representations. Our method leverages this technique for our density grid and canonical image as well, enabling efficient and rapid convergence of 3D modeling.

\noindent{\textbf{Neural scene editing.}}
Existing research on NeRF editing can be broadly categorized into two: one focuses on editing the geometry~\cite{chen2023neuraleditor,deng2021deformed,wang2021neus,yang2022neumesh,yuan2022nerfediting,zhang2023papr}; the other, known as style-based editing~\cite{haque2023instruct,pang2023locally,kobayashi2022decomposing,zhang2022arf,zhang2023refnpr}, aims to achieve scene stylization. Our research aligns with the latter category. 
Many NeRF stylization methods~\cite{zhang2022arf,huang2022consistent,phuoc2022stylized} have adopted techniques from 2D image stylization with style loss and content loss on images for NeRF optimization. While these methods can deliver 3D-consistent editing, they are primarily limited to global texture modifications and lack flexibility. Later, CLIP-NeRF~\cite{wang2022clipnerf} incorporates text conditions by regularizing the CLIP embeddings of the global scene with input prompts. Subsequent studies~\cite{kobayashi2022decomposing} extract 2D features such as DINO~\cite{caron2021emerging,oquab2023dinov2} for local editing; IN2N~\cite{haque2023instruct} proposes an iterative approach to edit the input images using pre-trained diffusion models~\cite{brooks2023instructpix2pix} for underlying NeRF optimization. Despite achieving high-fidelity editing results, most NeRF-based methods~\cite{chen2024mvip-nerf,dihlmann2024signerf,he2024customize,he2024freditor,jung2024geometry,khalid2024latenteditor,koo2024posterior,liu2024genn2n,mazzucchelli2024irene,mirzaei2024watch,radl2024laenerf,rojas2024datenerf} necessitate optimization for each text prompt or reference image, which can be inefficient in practical use. More recently, many works~\cite{chen2024gaussianeditor,fang2024gaussianeditor,qiu2024feature,wang2024gscream,wang2024view,wu2024gaussctrl,xu2024texture,ye2024gaussian,jaganathan2024iceg} start explore 3D editing with Gaussian Splatting~\cite{kerbl2023gs}. , As we generally focus on NeRF-based editing alternatives, we choose GaussianEditor~\cite{chen2024gaussianeditor} among them as the representative comparing baseline.

\noindent{\textbf{Neural atlases.}}
Our work shares similar insights with the research area of neural atlases~\cite{kasten2021layered,ye2022deformable,ouyang2023codef,ma2022nep}, which decompose videos into a canonical form with learned deformations, thereby enabling consistent video editing. Approaches such as neural layered atlases~\cite{kasten2021layered} employ an implicit network to distinguish foreground and background movement, dividing them into distinct layers. The CoDeF~\cite{ouyang2023codef} methodology represents 2D videos using content deformation fields by integrating 3D hash tables~\cite{muller2022instantngp}. However, these approaches lack 3D priors, limiting the effectiveness of 3D viewpoint changes in 3D scene editing.
\begin{figure*}[t]
    \centering
    \includegraphics[width=0.90\linewidth]{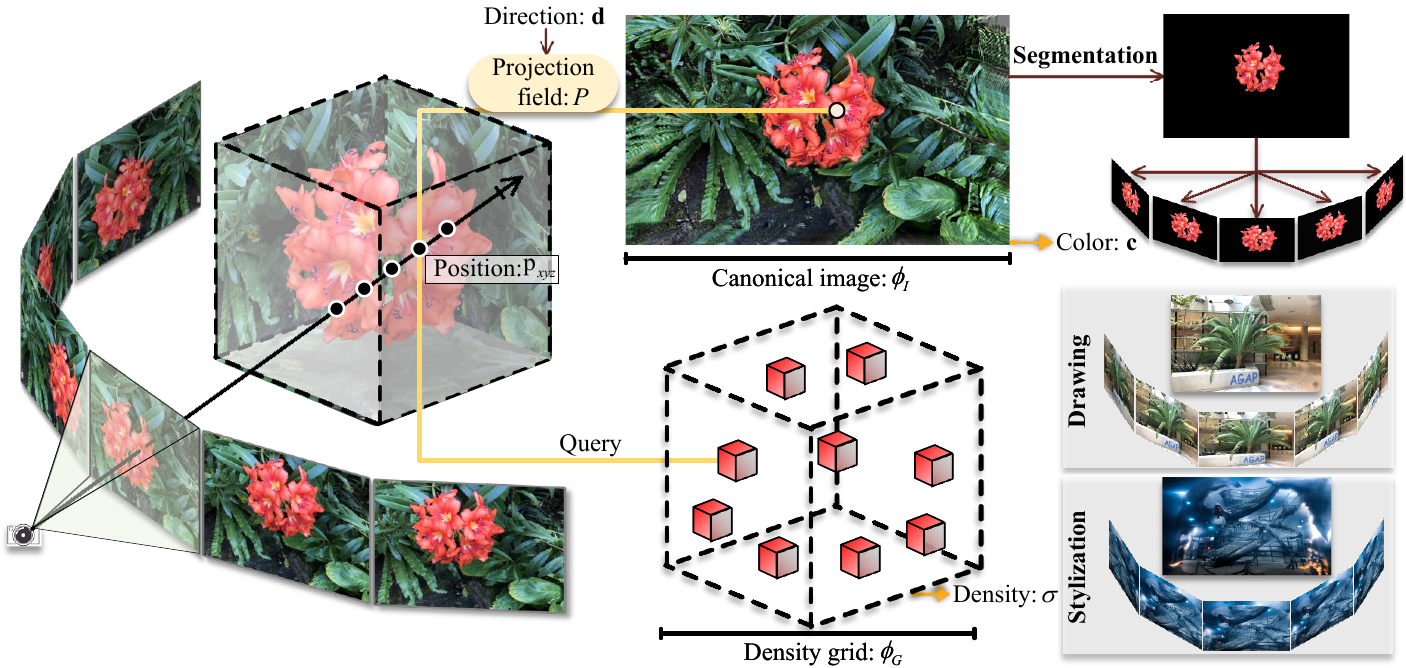}
    \centering
    \vspace{-5pt}
    \caption{
        \textbf{The overall pipeline}. \method consists of two components: (1) an explicit 3D density grid $\phi_G$ to estimate geometry for density $\sigma$; (2) an explicit canonical image $\phi_I$ with an associated view-dependent projection field $P$ to aggregate appearance for color $\mathbf{c}$. By performing 2D image processing on the canonical image, our method enables various editing (\textit{e.g.}, instance segmentation, interactive drawing, and scene stylization) through volume rendering without the need for re-optimization.
    }
    \label{fig:overview}
    \vspace{-10pt}
\end{figure*}

\section{Method}\label{sec:method}

As shown in~\cref{tab:functionality}, existing editing methods~\cite{zhang2022arf,zhang2023refnpr,haque2023instruct,chen2024gaussianeditor} necessitate several minutes or even hours per edit to re-optimize their original models.
Meanwhile, we argue that their implicit editing procedures through re-optimization reduce the level of user interactivity compared to explicit editing ways.
In view of such deficiencies, the core concept of \method is to learn a 2D canonical image as the interactive medium and allow users to do efficient and explicit 3D editing by lifting image processing on the 2D canonical image.
Under such an idea, the key challenges involve designing a projection field that bridges the appearance of the 3D scene appearance with the 2D canonical image and ensuring the naturalness of the canonical image.

\subsection{Preliminary}
Volume rendering~\cite{kajiya1984ray,mildenhall2020nerf} accumulates colors and densities of the 3D points sampled along the camera rays to render images. For a given camera ray $\mathbf{r}(t) = \mathbf{o} + t\mathbf{d}$ denoted by its origin $\mathbf{o} \in \mathbb{R}^{3}$ and direction $\mathbf{d} \in \mathbb{R}^{3}$, we sample $N$ points $\{\mathbf{r}(t_i)\}_{i=1}^{N}$ along the ray defined by a sorted distance vector $\mathbf{t} = [t_1, ..., t_N]^T \in \mathbb{R}^N$. 

NeRF~\cite{mildenhall2020nerf} models the 3D scene implicitly and leverage MLP networks to decode the density $\sigma_i = \mathtt{MLP}(\mathbf{r}(t_i))$ and the view-dependent color $\mathbf{c}_i = \mathtt{MLP}(\mathbf{r}(t_i), \mathbf{d})$ of a point located at $\mathbf{r}(t_i)$ on the ray with viewing direction $\mathbf{d}$. To render the image pixel $\hat{C}(\mathbf{r})$, we apply discretized volume rendering by Max~\cite{max1995optical} along the $N$ sampled ray points with $\delta_i$ denoting the distance to the nearby sampled points:
\begin{equation}
\begin{split}
    \hat{C}(\mathbf{r}) &= \sum_{i=1}^{N} {T_i (1 - \exp(-\sigma_i\delta_i))\mathbf{c}_i}, \text{where}\\
    T_i &= \exp(-\sum_{j=1}^{i-1} {\sigma_j\delta_j}).
\end{split}
\end{equation}
Training such MLPs for 3D radiance field modeling requires observed images with known camera poses. Specifically, NeRF model is optimized by minimizing the average $\mathcal{L}_2$ distance between the rendered pixel color $\hat{C}(\mathbf{r})$ and the ground-truth pixel color $C(\mathbf{r})$:
\begin{equation}
    \mathcal{L}_{color} = \frac{1}{|\mathcal{R}|} \sum_{\mathbf{r} \in \mathcal{R}} {\left\|\hat{C}(\mathbf{r}) - C(\mathbf{r}) \right\|_2^2}.
\end{equation}

\subsection{Model Formulation}
% Our representation disentangles the density and color of the scene and uses different modalities for modeling. 
Formally, given a set of multi-view training images $\mathcal{I}$, \method models the scene appearance by an explicit canonical image $\phi_I$ plus a corresponding implicit projection field $P$ inspired by~\cite{ouyang2023codef,park2021hypernerf,park2021nerfies}; the scene geometry is estimated by an explicit 3D density grid $\phi_G$. 
With such a representation, one can render different views through volume rendering. 
Our key property is that by explicitly editing the canonical image $\phi_I$, it can propagate the edited appearance to the whole scene through the projection field $P$ without any re-optimization. Our overall framework is shown in~\cref{fig:overview}.

\noindent\textbf{Density.} 
In 3D modeling, textures are applied to the mesh surface to provide visual details such as colors, patterns, and material properties~\cite{wang2018pixel2mesh,ji2017surfacenet}.
We opt for an explicit voxel-grid representation~\cite{sun2022dvgo,fridovichkeil2022plenoxels,chen2022tensorf,muller2022instantngp} instead of an implicit MLP such as NeRF~\cite{mildenhall2020nerf} to achieve fast convergence and efficient query.
Given a particular query point $\mathbf{p}_{xyz} \triangleq \mathbf{r}(t_i) \in \mathbb{R}^3$, we obtain the corresponding density $\sigma \in \mathbb{R}$ via a trilinear interpolation, followed by a Softplus activation:
\begin{equation}
    \sigma = \mathtt{Softplus} (\mathtt{GridSample} (\mathbf{p}_{xyz}, \phi_G)),
\end{equation}
where $\phi_G$ denotes the one-channel voxel grid with learnable parameter $\phi_g$ at a voxel resolution size of $N_x \times N_y \times N_z$. 
With a density grid with explicit parameterization, we hope our model can obtain a coarsely accurate density estimation of the 3D geometry at the early training stage, facilitating the learning of 2D appearance aggregation. 
Such a choice is also proven to be crucial by our experiments in~\cref{subsec:ablation}. 

\noindent\textbf{Appearance.} 
In order to empower efficient 3D editing capabilities, we formulate the color appearance by an explicit canonical image $\phi_I \in \mathbb{R}^{H \times W \times 3}$ with an associated view-dependent implicit projection field $P (\cdot, \cdot): (\mathbb{R}^3, \mathbb{R}^3) \to \mathbb{R}^2$, where $H$ and $W$ represent the image height and width, respectively. 
This formulation maps a given query point $\mathbf{p}_{xyz}$ in the 3D field with viewing direction $\mathbf{d}$ to the projected 2D point $\mathbf{p}_{uv}$ on the canonical image $\phi_I$. The projection point $\mathbf{p}_{uv}$ is then used to query the RGB color $\mathbf{c} \in \mathbb{R}^{3}$ from the canonical image $\phi_I$ via interpolation:
\begin{equation}
    \mathbf{c} = \mathtt{Sigmoid} (\mathtt{GridSample} (\mathbf{p}_{uv}, \phi_I)).
\end{equation}

\subsection{Canonical Projection with Projection Offset}
% The design of the projection field is crucial to achieving naturalness and completeness in the learned canonical image. 
%
We model the projection field $P$ as a non-learnable canonical projection $P_c$ with a projection offset learning $P_o$.
To model different 3D scenes, we choose suitable foundational projection as the canonical projection $P_c$ derived from different real camera models, which plays a significant role in ensuring naturalness and completeness in the learned canonical.
The projection offset $P_o$ aims to address view-dependent effects and handle occlusions in complex scenes. 

Specifically, the canonical projection $P_c(\cdot): \mathbb{R}^3 \to \mathbb{R}^2$ projects the query 3D point $\mathbf{p}_{xyz}$ to an initial 2D projection $\tilde{\mathbf{p}}_{uv} = P_c(\mathbf{p}_{xyz})$ on the canonical. The projection offset $\Delta\mathbf{p}_{uv}$ is modeled by $P_o$ with parameter weights $\phi_P$:
\begin{equation}
\label{equ:projection_offset}
    \Delta\mathbf{p}_{uv} = P_o(\mathbf{p}_{xyz}, \mathbf{d}; \phi_P).
\end{equation}
For simplicity, we omit the 3D positional encoding $\gamma_{p}$ of $\mathbf{p}_{xyz}$ and the viewing direction encoding $\gamma_{d}$ of $\mathbf{d}$ in~\cref{equ:projection_offset}.

The final projection point $\mathbf{p}_{uv}$ is formulated as follows:
\begin{equation}
    \mathbf{p}_{uv} = \tilde{\mathbf{p}}_{uv} + \Delta\mathbf{p}_{uv}.
\end{equation}

\subsection{Optimization and Regularization}

\noindent\textbf{Projection regularization.} 
In order to obtain a visually natural canonical image $\phi_I$, one important regularization is to avoid the deviation from the perception by the defined pseudo canonical camera. 
We find the following simple regularization works well and stabilizes the training:
\begin{equation}
    \mathcal{L}_{uv} = \left\| \Delta\mathbf{p}_{uv} \right\|_2^2.
\end{equation}

\noindent\textbf{Total variation regularization.} To mitigate floating densities, we incorporate total variation regularization~\cite{rudin1992nonlinear} $\mathcal{L}_{tv}$ into the density grid $\phi_G$. This regularization term is particularly beneficial during the initial stages of training.

\noindent\textbf{Optimization objective.} 
% The final optimization process of our method to model the scene for efficient 3D editing is formulated as follows:
The final optimization process of our method is formulated as follows:
\begin{equation}
    \phi_G^*, \phi_I^*, \phi_P^* = \argmin_{\phi_G, \phi_I, \phi_P}{\mathcal{L}_{color} + \mathcal{L}_{uv} + \mathcal{L}_{tv}}.
\end{equation}

\begin{figure*}[t]
    \small
    \begin{center}
        \begin{subfigure}{1\linewidth}
            \includegraphics[width=1\linewidth]{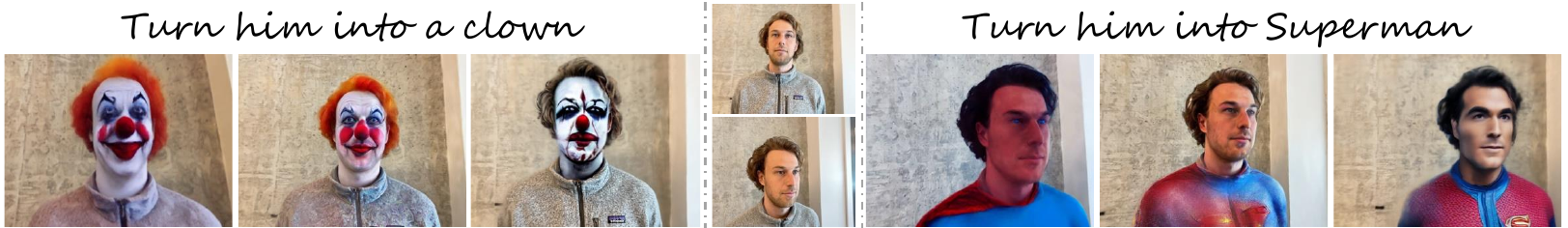}
        \end{subfigure}
        \begin{subfigure}{1\linewidth}
            \begin{subfigure}{0.145\linewidth}\hfil IN2N~\cite{haque2023instruct}                \end{subfigure}
            \begin{subfigure}{0.155\linewidth}\hfil GaussianEditor~\cite{chen2024gaussianeditor} \end{subfigure}
            \begin{subfigure}{0.145\linewidth}\hfil Ours                                         \end{subfigure}
            \begin{subfigure}{0.085\linewidth}\hfil Refs                                         \end{subfigure}
            \begin{subfigure}{0.145\linewidth}\hfil IN2N~\cite{haque2023instruct}                \end{subfigure}
            \begin{subfigure}{0.155\linewidth}\hfil GaussianEditor~\cite{chen2024gaussianeditor} \end{subfigure}
            \begin{subfigure}{0.145\linewidth}\hfil Ours                                         \end{subfigure}
        \end{subfigure}
        \begin{subfigure}{1\linewidth}
            \includegraphics[width=1\linewidth]{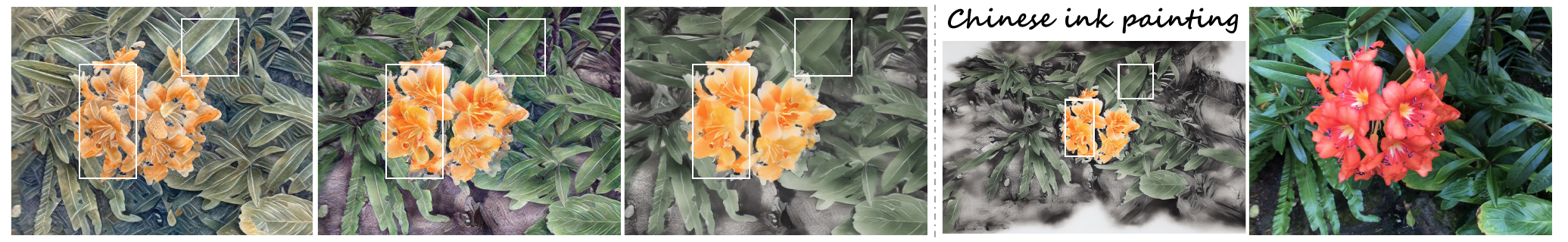}
        \end{subfigure}
        \begin{subfigure}{1\linewidth}
            \begin{subfigure}{0.195\linewidth}\hfil ARF~\cite{zhang2022arf}        \end{subfigure}
            \begin{subfigure}{0.195\linewidth}\hfil Ref-NPR~\cite{zhang2023refnpr} \end{subfigure}
            \begin{subfigure}{0.195\linewidth}\hfil Ours                           \end{subfigure}
            \begin{subfigure}{0.195\linewidth}\hfil Reference Style             \end{subfigure}
            \begin{subfigure}{0.195\linewidth}\hfil Reference View                 \end{subfigure}
        \end{subfigure}
    \end{center}
    \vspace{-18pt}
    \caption{
        \textbf{Visual comparison of novel-view scene stylization results} on the \textit{IN2N} and \textit{LLFF} dataset given different text prompts or image reference. Our method can achieve on-par stylization results with the baselines while requiring no time-consuming re-optimization procedures. As highlighted in row two, our method can better preserve color and textural consistencies aligning with the image reference.
    }
    \label{fig:main_stylization1}
    \vspace{-10pt}
\end{figure*}

\begin{figure}[t]
    \includegraphics[width=1\linewidth]{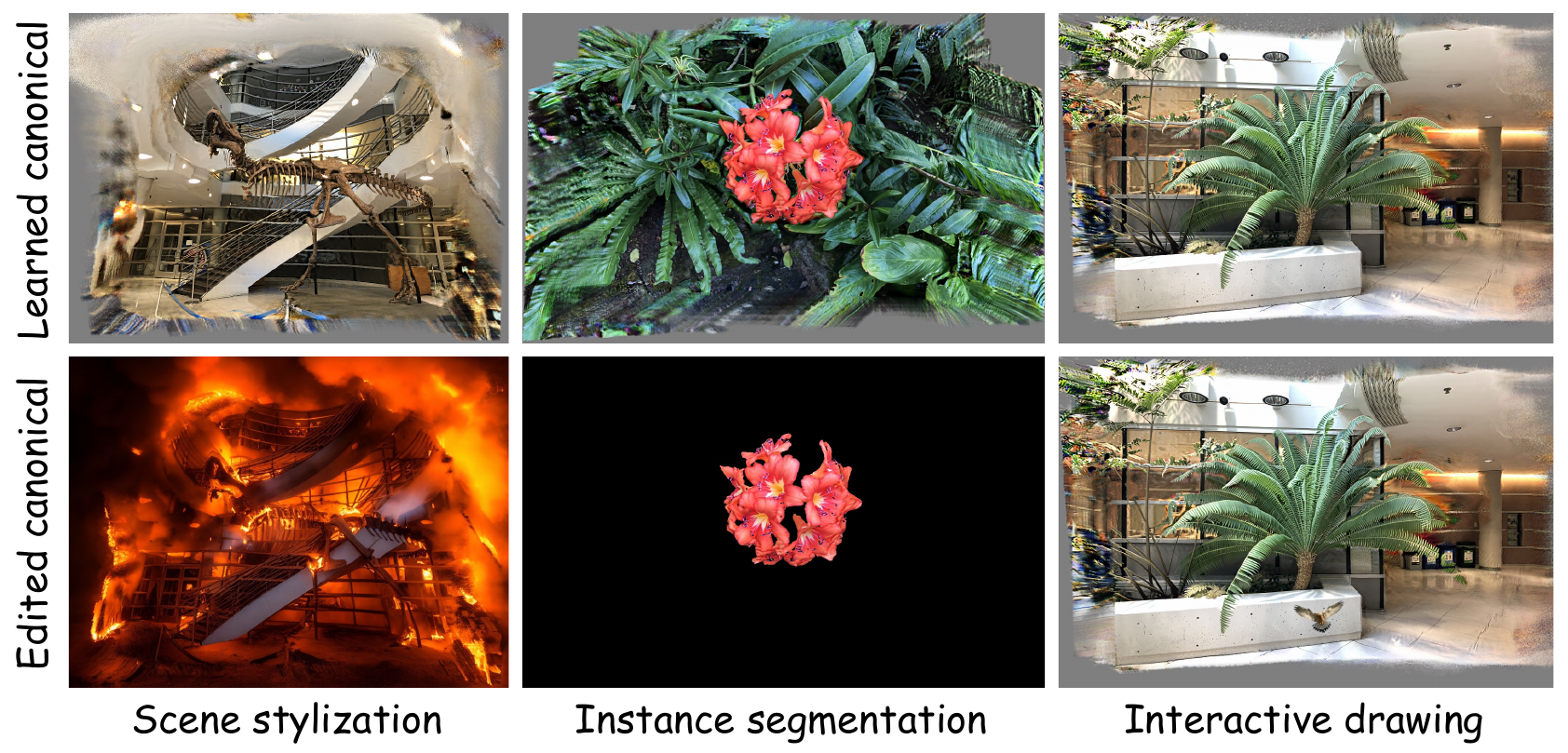}
    \vspace{-8pt}
    \caption{
        By performing explicit edits on the canonical image $\phi_I$, our model propagates the editing effects through the learned projection field $P$ for efficient 3D editing.
    }
    \label{fig:editing}
    \vspace{-5pt}
\end{figure}

\section{Experiments}\label{sec:exp}

Recall that the canonical projection $P_c$ can be viewed as positioning a pseudo-canonical camera within the 3D scene. This the crucial step in achieving a natural-looking canonical image $\phi_I$ by initializing the projection field $P$, by choosing a suitable foundational projection derived from different real camera models for the canonical projection $P_c$ when modeling different types of 3D data scenes.

\subsection{Data Types}
\label{subsec:data-types}

\noindent\textbf{Forward-facing data.} 
The \textit{LLFF}~\cite{mildenhall2019llff} dataset comprises real-world forward-facing scenes, where each scene is accompanied by several training images captured by handheld cameras placed in a rough grid pattern nearly on a vertical plane. We evaluate the dataset at a resolution of $756 \times 1008$.

We utilize the normalized device coordinate (NDC) space to model the forward-facing captures, where the pseudo canonical camera is defined as the average camera pose at the world origin. 
The perspective projection $f_p$ can map any given point $\mathbf{p}_{xyz}$ in the world coordinates to its corresponding NDC point $f_p(\mathbf{p}_{xyz}) \triangleq (p_{x'}, p_{y'}, p_{z'})$.
The coordinate range of the voxel grid $\phi_G$ is defined by the bounding box in NDC space. 
One notable property of NDC is that for a ray $\mathbf{r}$ origin from the camera center, all points along the ray share identical values for $p_{x'}$ and $p_{y'}$. 
Leveraging this property, we can design an appropriate choice for the canonical projection: $\tilde{\mathbf{p}}_{uv} \triangleq (p_{x'}, p_{y'})$.

\noindent\textbf{Panorama data.} 
The \textit{Replica}~\cite{straub2019replica} dataset is a collection of various high-quality and high-resolution 3D reconstructions of indoor scenes with clean and dense geometry. We evaluate the panorama scenes as processed in SOMSI~\cite{habtegebrial2022somsi}, where each scene is rendered as a grid of equally spaced spherical images at a resolution of $1024 \times 512$.

The panorama data captures outward-facing views covering a 360-degree field of view around the global center. 
We place the pseudo-canonical camera at the global origin and define the Equirectangular projection as the canonical projection.
Drawing inspiration from the smooth coordinate transforms in~\cite{barron2022mipnerf360,reiser2022merf}, we introduce a new contracted formulation $f_c$ specifically for panorama scenes:
\begin{equation}
    f_c(\mathbf{x}) = \frac{\mathbf{x}}{\left\| \mathbf{x} \right\|} (1 - \frac{1}{\left\| \mathbf{x} \right\| + 1}),
\end{equation}
where $\mathbf{x} \in \mathbb{R}^3$ is a 3D point in Euclidean space. 
This design transforms the points in such a way that they are distributed proportionally to the disparity in a unit sphere. 
Accordingly, the voxel grid $\phi_G$ is defined as a cube with range $[-1, 1]^3$.
Given $f_c(\mathbf{p}_{xyz}) \triangleq (p_{x'}, p_{y'}, p_{z'})$, the canonical projection $P_c$ from a 3D point $\mathbf{p}_{xyz}$ to a 2D canonical image pixel $\tilde{\mathbf{p}}_{uv} \triangleq (\tilde{p}_u, \tilde{p}_v)$ can be formulated as follows:
\begin{equation}
\label{equ:contracted}
\begin{alignedat}{3}
    \tilde{p}_{u} &= \tan^{-1} (\frac{p_{y'}}{p_{x'}}) &\in [-\pi, \pi], \\
    \tilde{p}_{v} &= \sin^{-1} (p_{z'}) &\in [-\frac{\pi}{2}, \frac{\pi}{2}].
\end{alignedat}
\end{equation}

\begin{figure*}[t]
    \begin{subfigure}{0.015\linewidth}
        \small
        \parbox[][0.0cm][c]{\linewidth}{
            \raisebox{1.9in}{\centering\rotatebox{90}{Novel Views \quad\quad Canonical Image}}
        }
    \end{subfigure}
    \begin{subfigure}{0.985\linewidth}
        \includegraphics[width=1\linewidth]{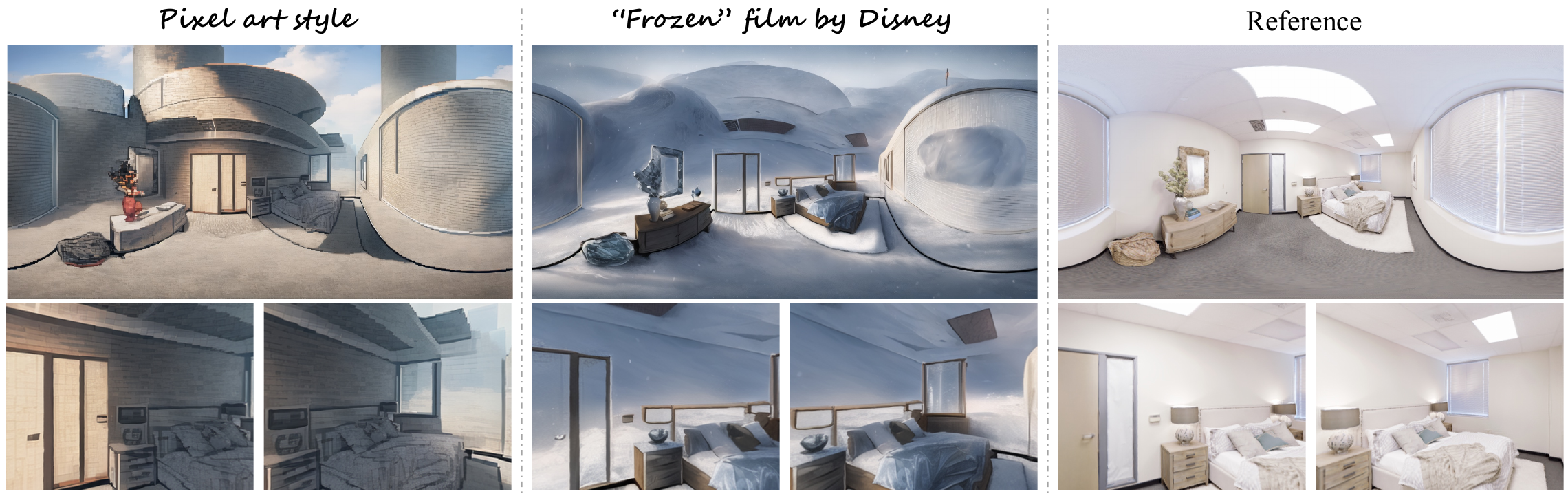}
    \end{subfigure}
    \vspace{-18pt}
    \caption{
        \textbf{More visualization of scene stylization results} on the panorama \textit{Replica} dataset given different text prompts.
    }
    \label{fig:main_stylization3}
    \vspace{-8pt}
\end{figure*}

\begin{figure}[t]
    \small
    \centering
    \begin{subfigure}{0.04\linewidth}
        \small
        \parbox[][0.0cm][c]{\linewidth}{\centering\raisebox{1.5in}{\rotatebox{90}{Background \quad Foreground}}}
    \end{subfigure}
    \begin{subfigure}{0.95\linewidth}
        \includegraphics[width=1.0\linewidth]{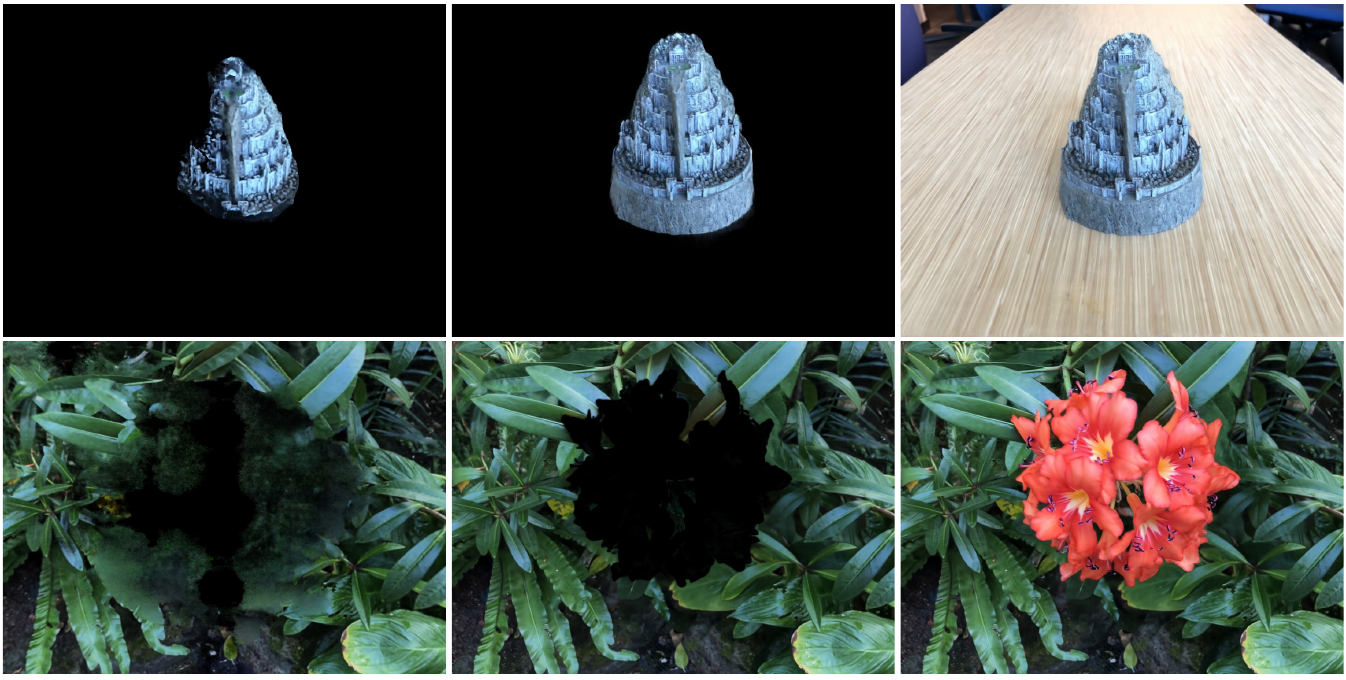}
    \end{subfigure}
    % \begin{subfigure}{0.99\linewidth}\ \end{subfigure}
    \begin{subfigure}{0.04\linewidth}\ \end{subfigure}
    \begin{subfigure}{0.95\linewidth}
        \begin{subfigure}{0.328\linewidth}\hfil DFFs~\cite{kobayashi2022decomposing}\end{subfigure}
        \begin{subfigure}{0.328\linewidth}\hfil Ours\end{subfigure}
        \begin{subfigure}{0.328\linewidth}\hfil Reference\end{subfigure}
    \end{subfigure}
    \caption{
        \textbf{Visual comparison of foreground and background segmentation results} on the \textit{LLFF} dataset.
    }
    \label{fig:main_segmentation1}
    \vspace{-8pt}
\end{figure}

\noindent\textbf{Object-centric data.}
The \textit{NeRF-Synthetic}~\cite{mildenhall2020nerf} dataset comprises synthetic objects with intricate geometry and realistic materials. Each object in the dataset includes 100 training views and 200 test views, all rendered at a resolution of $800 \times 800$.
The \textit{DTU}~\cite{jensen2014dtu} dataset is formulated in a object-level forward-facing setting, where training images are on quarter hemispheres ($\sim \frac{1}{8}$ spheres) with object masks. We evaluate this dataset at a resolution of $600 \times 800$ after downscaling the images by a factor of 2.
The \textit{NeUVF}~\cite{ma2022nep} dataset capture human head videos using 12 calibrated cameras located at a hemisphere with approximately 120$^{\circ}$. We use their first frames for static scenes, to specifically to model and edit the scenes that prominently feature human subjects for qualitative evaluation.

Analogous to Earth map, we can place the pseudo canonical camera at the global origin to model object-centric data using a (partial hemispheric) Equirectangular projection. 
Similar to~\cref{equ:contracted}, the canonical projection $P_c$ from a 3D point $\mathbf{p}_{xyz} \triangleq (p_{x}, p_{y}, p_{z})$ to a 2D canonical image pixel $\tilde{\mathbf{p}}_{uv} \triangleq (\tilde{p}_u, \tilde{p}_v) = (\tan^{-1} (\tfrac{p_{y}}{p_{x}}), \sin^{-1} (p_{z}))$.

\noindent\textbf{Unbounded 360-degree data.}
The \textit{Mip-NeRF 360}~\cite{barron2022mipnerf360} and the \textit{IN2N}~\cite{haque2023instruct} dataset comprise unbounded outdoor and indoor scenes, which is the most challenging case. 
Each scene showcases a complex background along with a central object or area, captured at varying high resolutions.

Learning a good 3D-to-2D projection field $P$ for unbounded 360-degree scenes is indeed a non-trivial research challenge~\cite{desbrun2002intrinsic}. 
We leverage two canonical images for modeling the foreground central objects using (hemispheric) Equirectangular projection and the unbounded background by the contracted formulation as panorama, respectively.

\begin{figure*}[t]
    \small
    \begin{center}
        \begin{subfigure}{1\linewidth}
            \includegraphics[width=1\linewidth]{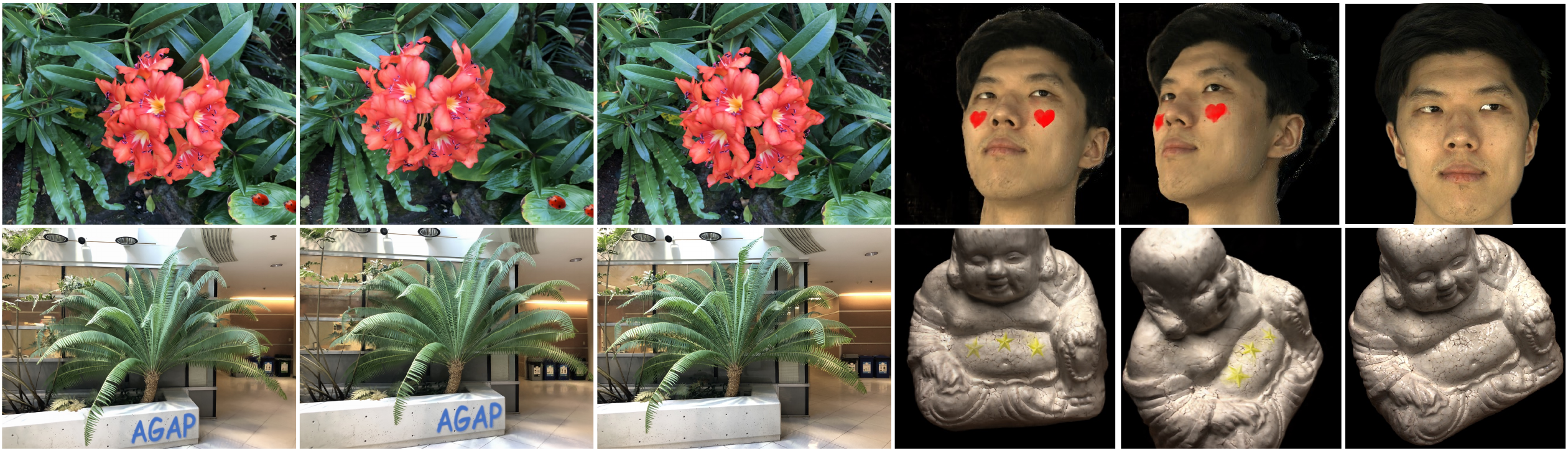}
        \end{subfigure}
        \begin{subfigure}{1\linewidth}
            \begin{subfigure}{0.185\linewidth}\hfil Edited NVS 1 \end{subfigure}
            \begin{subfigure}{0.185\linewidth}\hfil Edited NVS 2 \end{subfigure}
            \begin{subfigure}{0.185\linewidth}\hfil Reference    \end{subfigure}
            \begin{subfigure}{0.138\linewidth}\hfil Edited NVS 1 \end{subfigure}
            \begin{subfigure}{0.138\linewidth}\hfil Edited NVS 2 \end{subfigure}
            \begin{subfigure}{0.138\linewidth}\hfil Reference    \end{subfigure}
        \end{subfigure}
    \end{center}
    \vspace{-12pt}
    \caption{
        \textbf{Visualization of texture editing (\textit{i.e.}, drawing) results} at different novel viewpoints on the \textit{LLFF}, \textit{NeUVF}, and \textit{DTU} dataset.
    }
    \label{fig:main_drawing}
    \vspace{-5pt}
\end{figure*}

\begin{figure*}
    \centering
    \includegraphics[width=0.95\linewidth]{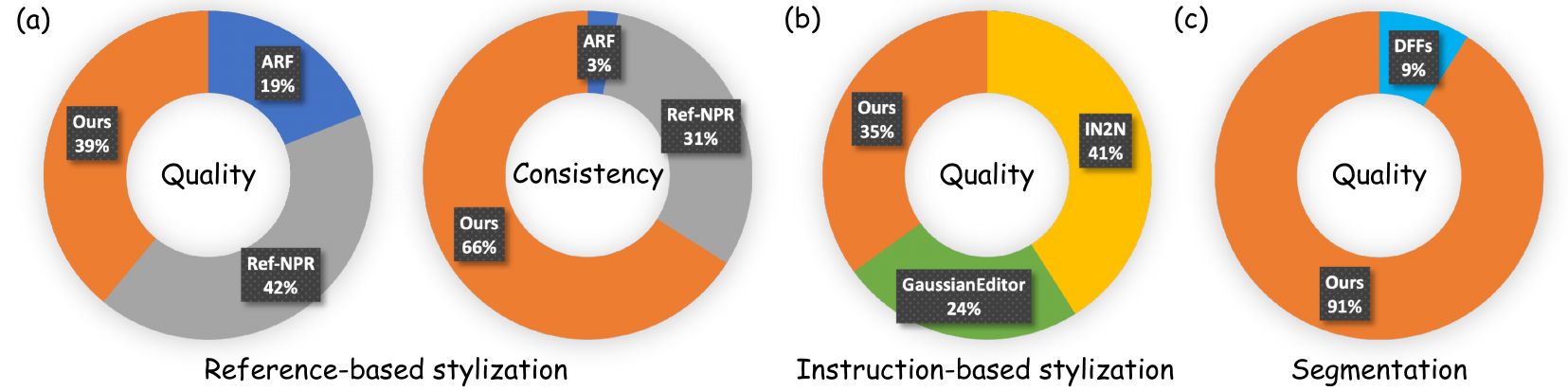}
    \vspace{-8pt}
    \caption{A user study of our method with different alternatives in terms of both perceptual editing quality and consistency.}
    \vspace{-12pt}
    \label{fig:user-study}
\end{figure*}

\subsection{Implementation} 

Our pipeline involves a two-step process: 1) training a per-scene reconstruction model; 2) subsequent explicit edits on the canonical image $\phi_I$ for efficient 3D scene editing.

\noindent\textbf{Training details.} 
By default, we optimize a per-scene model for $60$k steps using the Adam optimizer~\cite{kingma2015adam} with an initial learning rate of $0.1$ for both the explicit density grid $\phi_G$ and the canonical image $\phi_I$, and a learning rate of $0.001$ for the implicit projection field $P$ with learnable parameters $\phi_P$. All experiments are conducted and tested on a single RTX A6000 GPU. For other implementation details and hyperparameters, please see the supplementary materials.

\noindent\textbf{Editing pipeline.}
After we obtain the pre-trained model using our novel 3D representation, users can perform explicit edits as shown in~\cref{fig:editing} on the canonical image $\phi_I$ for various efficient 3D scene editing functionalities, such as scene stylization, instance segmentation, and texture editing. Our model can propagating the editing effects through the learned projection field $P$. Specifically, we utilize the prompt-guided \textit{ControlNet}~\cite{zhang2023controlnet} for scene stylization and \textit{Segment-anything (SAM)}~\cite{kirillov2023sam} for instance segmentation. As for texture editing, users are free to draw or write on the canonical image.

\begin{figure*}[t]
    \begin{subfigure}{0.99\linewidth}
        \includegraphics[width=1\linewidth]{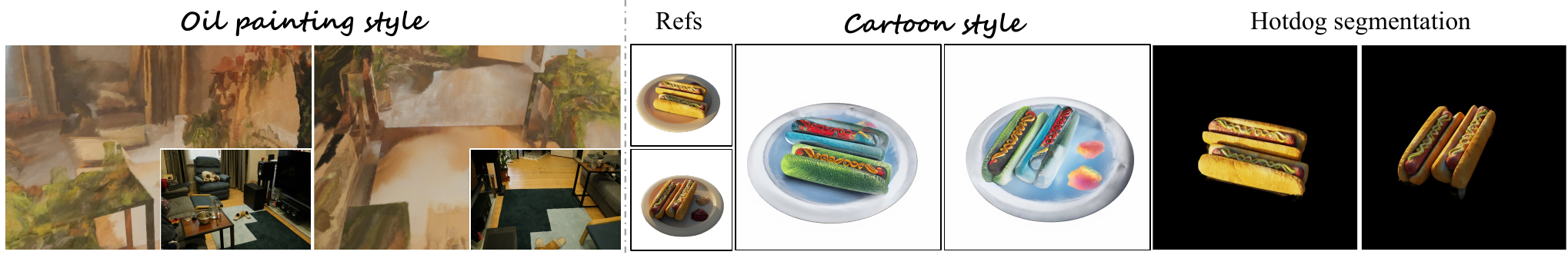}
    \end{subfigure}
    \vspace{-8pt}
    \caption{
        \textbf{More visualization of editing results} on the 360$^{\circ}$ and object-centric datasets.
    }
    \label{fig:main_stylization4}
    \vspace{-5pt}
\end{figure*}

\begin{table}[t]
    \small
    \centering
    \caption{
        \textbf{Reconstructed PSNR} on the \textit{LLFF} and \textit{Replica} datasets.
    }
    \label{tab:comparison}
    \vspace{-8pt}
    \SetTblrInner{rowsep=1.2pt}      % Row space.
    \SetTblrInner{colsep=8.0pt}      % Col space.
    \begin{tblr}{
        cells={halign=c,valign=m},   % Text alignment for all cells.
        column{1}={halign=l},        % Text alignment for the first column.
        hline{1,2,6,8}={1-3}{},        % Horizontal lines.
        hline{1,8}={1.0pt},          % Horizontal line width.
        vline{2}={1-8}{},            % Vertical lines.
    }
    Methods                            & \textit{LLFF} dataset & \textit{Replica} dataset \\
    LLFF~\cite{mildenhall2019llff}     & 24.13                 & -                        \\
    NeRF~\cite{mildenhall2020nerf}     & 26.50                 & -                        \\
    DVGO~\cite{sun2022dvgo}            & 26.34                 & -                        \\
    SOMSI~\cite{habtegebrial2022somsi} & -                     & 39.54                    \\
    Ours (PE)                          & 24.83                 & 38.42                    \\
    Ours (Hash)                        & 26.20                 & 38.68                    \\
    \end{tblr}
    \vspace{-13pt}
\end{table}

\subsection{Evaluation on Editability}

\noindent\textbf{Scene stylization.} We conduct a comparative analysis between our method and three state-of-the-art stylization methods: ARF~\cite{zhang2022arf}, Ref-NPR~\cite{zhang2023refnpr}, and IN2N~\cite{haque2023instruct}. Specifically, ARF and Ref-NPR rely on one or multiple stylized reference images, while IN2N utilizes text prompts through Diffusion models~\cite{brooks2023instructpix2pix} for guidance. All these baseline methods require additional optimization processes to achieve stylization given a specific style, while our method is optimization-free. As shown in~\cref{tab:functionality}, our editing speed for stylization is approximately 20$\times$ faster than ARF and Ref-NPR and approximately 500$\times$ faster than IN2N. 

\begin{table}[t]
    \small
    \centering
    \caption{
        PSNR \textbf{ablations} of model components on \textit{trex} scene.
    }
    \label{tab:ablation}
    \vspace{-8pt}
    \SetTblrInner{rowsep=1.2pt}      % Row space.
    \SetTblrInner{colsep=5.3pt}      % Col space.
    \begin{tblr}{
        cells={halign=c,valign=m},   % Text alignment for all cells.
        column{1}={halign=l},        % Text alignment for the first column.
        hline{1,2,5,7}={1-4}{},        % Horizontal lines.
        hline{1,6}={1.0pt},          % Horizontal line width.
        vline{2}={1-5}{},            % Vertical lines.
    }
    Settings                             &  PE     &  Hash   \\
    \uppercase\expandafter{\romannumeral1}. No canonical projection $P_c$           &  23.18  &  27.56  \\
    \uppercase\expandafter{\romannumeral2}. No projection offset $P_o$                   &  23.81  &  23.88  \\
    \uppercase\expandafter{\romannumeral3}. No viewdir $\mathbf{d}$ in projection offset $P_o$        &  25.50  &  26.76  \\
    Full model                           &  25.85  &  27.24  \\
    \end{tblr}
    \vspace{-5pt}
\end{table}

In~\cref{fig:main_stylization1}, we demonstrate some comparing visualizations evaluated on the IN2N dataset and the LLFF dataset. 
In the first row, we present visual results from the IN2N dataset using different text prompts. While both our methods can effectively edit the scene to the desired style, the IN2N baseline necessitates approximately 3 hours to optimize the underlying NeRF model per edit, whereas our method requires no additional re-optimization. In the second row, we evaluate our method alongside the ARF and Ref-NPR baselines using the LLFF dataset. Both ARF and Ref-NPR produce implicit global stylization that looks like the reference style. However, our method achieves superior color and textural consistencies aligning with the provided reference style image. We present more visual results of stylization in~\cref{fig:main_stylization3} on the panorama dataset.

\noindent\textbf{Instance segmentation.} We evaluate our method with the state-of-the-art DFFs~\cite{kobayashi2022decomposing} method, which enables NeRFs to decompose a specific object given a text or image-patch query. The evaluation of baseline DFFs is only based on text query according to its official codebase. In~\cref{fig:main_segmentation1}, we show some comparing visualization of foreground and background segmentation. Our method allows users to do 3D instance segmentation easily by applying 2D segmentation of the desired objects in the canonical image.

\noindent\textbf{Texture editing.}
Our method can further do textural appearance editing of the 3D scene by drawing or painting on the canonical image. In~\cref{fig:main_drawing}, we can observe that our method ensures both appearance and 3D consistency in novel views. For the first sample in row one, we paint an ``AGAP'' logo on the marble pedestal of the fern plant; for the second sample in row two, we draw two ``ladybirds'' on the leaves at the right bottom. We further present more visualizations in~\cref{fig:main_stylization4} and the supplementary materials.

\noindent\textbf{User study.} 
A user study comparing perceptual quality with baseline methods is presented in~\cref{fig:user-study}. We expect our approach to achieve comparable  visual performance with existing alternatives, yet be far more efficient (at least 20$\times$ faster) and easier (\textit{e.g.}, supporting more types of editing) to use in practice as in~\cref{tab:functionality}.
% where we expect our approach to achieve comparable visual performance. 
The user study involves 43 participants, each answering 10 comparison questions against baseline methods to select the best option:
\begin{itemize}
    \item For \textbf{\textit{reference-based stylization}} (\textit{i.e.}, ARF and Ref-NPR), participants select the best aligned with the reference image in terms of editing quality and consistency.
    \item For \textbf{\textit{instruction-based stylization}} (\textit{i.e.}, IN2N and GaussianEditor), participants evaluate overall aesthetic appeal based on the provided instruction text prompt.
    \item For \textbf{\textit{segmentation}} comparisons (\textit{i.e.}, DFFs), participants determine whether ours or DFFs provides superior segmentation accuracy and editing performance.
\end{itemize}

\begin{figure}[t]
    \vspace{-10pt}
    \begin{subfigure}{0.293\linewidth} % 213
        \includegraphics[width=1\linewidth]{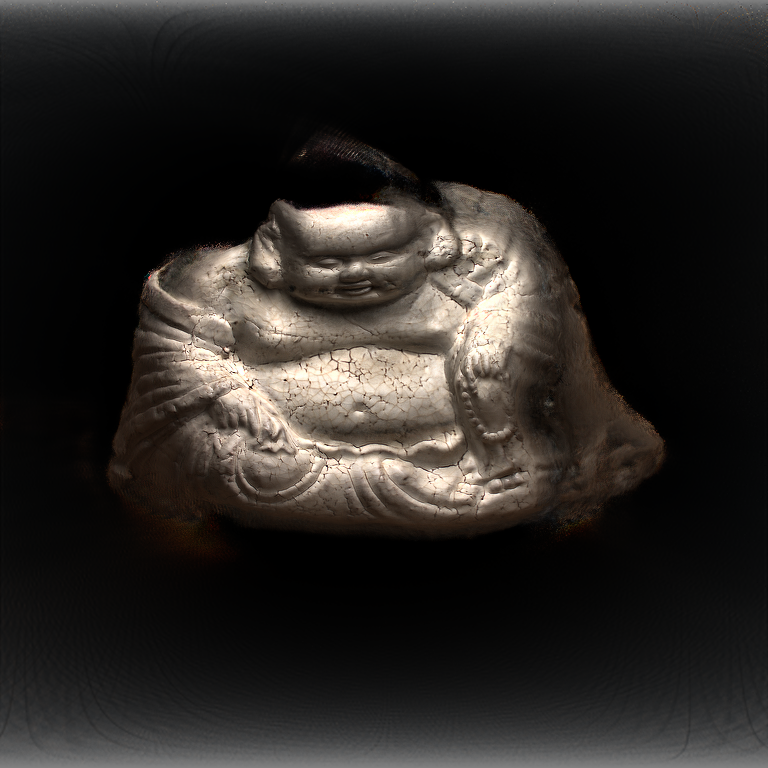}
        \subcaption{Ours (\textit{w/} $P_c$)}
        \label{subfig:canonical-ours}
    \end{subfigure}
    \begin{subfigure}{0.293\linewidth}
        \includegraphics[width=1\linewidth]{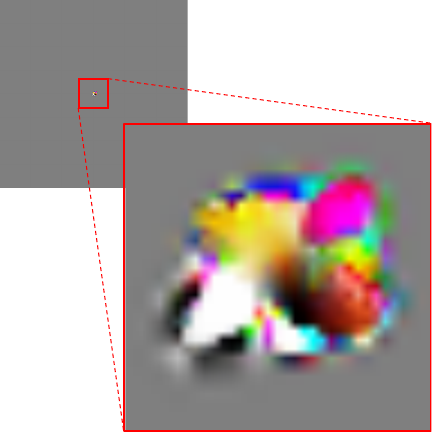}
        \subcaption{Ours (\textit{w/o} $P_c$)}
        \label{subfig:canonical-ablation}
    \end{subfigure}
    \begin{subfigure}{0.393\linewidth}
        \includegraphics[width=1\linewidth]{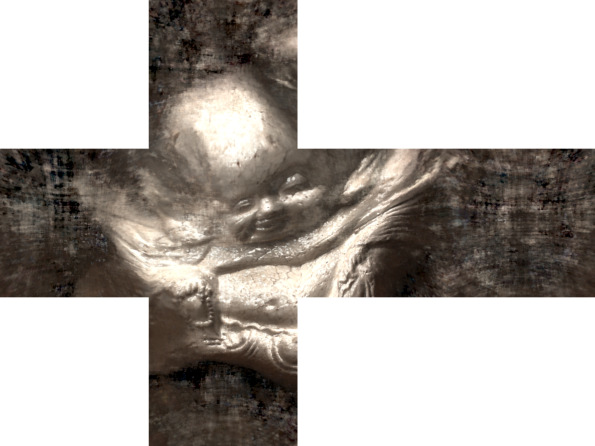}
        \subcaption{UV map~\cite{xiang2021neutex}}
        \label{subfig:canonical-uvmap}
    \end{subfigure}
    \vspace{-18pt}
    \caption{Canonical image of ablating different 2D projections.}
    \label{fig:ablation-canonical-projection}
    \vspace{-8pt}
\end{figure}

\begin{figure}[t]
    % \vspace{-10pt}
    \small
    \begin{subfigure}{1.0\linewidth}
        \includegraphics[width=1\linewidth]{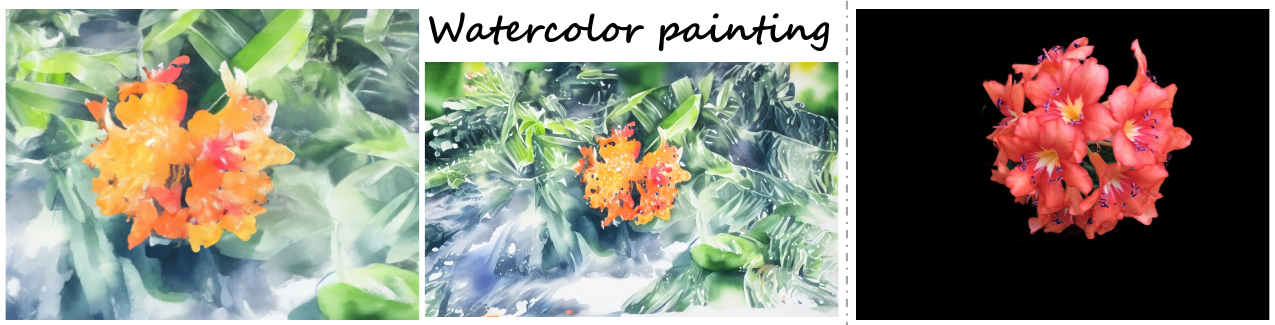}
    \end{subfigure}
    \begin{subfigure}{0.3\linewidth}\hfil Stylization NVS\end{subfigure}
    \begin{subfigure}{0.385\linewidth}\hfil Reference\end{subfigure}
    \begin{subfigure}{0.3\linewidth}\hfil Segmentation NVS\end{subfigure}
    \vspace{-13pt}
    \caption{Hash models showcase a moderate level of editability.}
    \label{fig:ablation-hash}
    \vspace{-13pt}
\end{figure}

\subsection{Ablation and Analysis.}
\label{subsec:ablation}

\noindent\textbf{Editability.}
When designing the projection field, we find that having a good canonical projection is essential for natural and user-friendly editing. As presented by Setting \uppercase\expandafter{\romannumeral1} in~\cref{tab:ablation}, where the canonical projection is removed it while keeping the others unchanged, we can see that the PE model experiences a drop while the hash model shows an increase. However, both models lose the ability to perform editing tasks as the learned canonical images $\phi_I$ deviate from natural images to latent color maps, where the visual comparison is shown in~\cref{subfig:canonical-ours,subfig:canonical-ablation}. Apart from the various options for suitable canonical projection when handling different data types in~\cref{subsec:data-types}, indeed, there exists some other projection functions like UV maps, but apparently they are hard to edit, as demonstrated in~\cref{subfig:canonical-uvmap}. 

\noindent\textbf{Reconstruction fidelity.}
\method, as a new efficient editing pipeline for 3D modeling, we also report the PSNR results here for completeness to showcase the capacity of faithful reconstruction in~\cref{tab:comparison}. 
We examine the capacity of our models with PE and Hash designs for the input $\mathbf{p}_{xyz}$ to learn the projection offset $P_o$. 
We find that PE models lead to superior editing capacity, whereas hash models prioritize reconstruction quality, meanwhile, maintaining a moderate level of editability as shown in~\cref{fig:ablation-hash}.

\begin{figure}[t]
    \vspace{-10pt}
    \small
    \begin{subfigure}{0.243\linewidth}
        \includegraphics[width=1\linewidth]{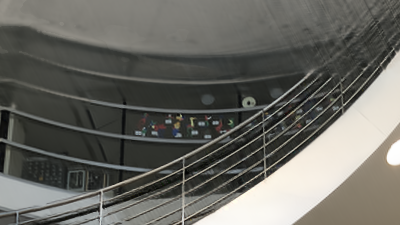}
    \end{subfigure}
    \begin{subfigure}{0.243\linewidth}
        \includegraphics[width=1\linewidth]{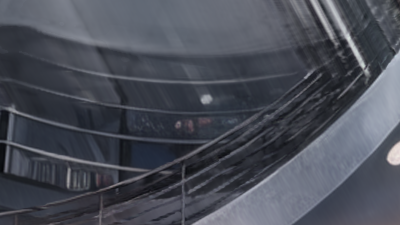}
    \end{subfigure}
    \begin{subfigure}{0.243\linewidth}
        \includegraphics[width=1\linewidth]{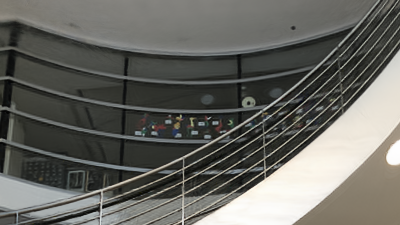}
    \end{subfigure}
    \begin{subfigure}{0.243\linewidth}
        \includegraphics[width=1\linewidth]{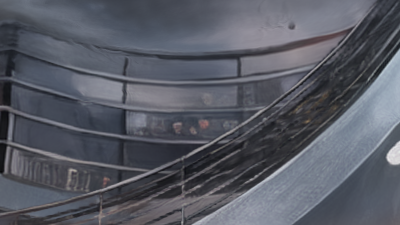}
    \end{subfigure}
    \begin{subfigure}{0.495\linewidth}\hfil Ours (\textit{w/o} offset $P_o$)\end{subfigure}
    \begin{subfigure}{0.495\linewidth}\hfil Ours (\textit{w/} offset $P_o$)\end{subfigure}
    \vspace{-13pt}
    \caption{Reconstruction and editing visual patches of ablating the projection offset $P_o$.}
    \label{fig:ablation-projection-offset}
    \vspace{-13pt}
\end{figure}

\noindent\textbf{Learnable projection offset.}
In Setting \uppercase\expandafter{\romannumeral2} in~\cref{tab:ablation}, where the entire projection offset $P_o$ is eliminated, a significant drop in performance is observed. In this case, the model cannot successfully handle the occlusion effects as presented in~\cref{fig:ablation-projection-offset} for both reconstruction and editing. In Setting \uppercase\expandafter{\romannumeral3} in~\cref{tab:ablation}, the removal of view-dependence from the learnable projection offset $P_o$ leads to a minor decrease in terms of statistical reconstruction, as the model no longer considers viewing directions.

\section{Discussion and Conclusion}\label{sec:conclusion}

This paper explores the task of efficient 3D scene editing, where we focus on editing efficiency and user interactivity. Specifically, we propose \method, an editing-friendly and efficient solution for neural 3D scene editing, by representing a 3D scene using a 2D canonical image equipped with a projection field, such that users can easily perform efficient 3D editing via processing the 2D image. The key challenge and contribution of our method lies in how to regularize the projection field to make the canonical image look natural.

Compared to existing baselines, which require a time-consuming optimization process for one editing style, our approach can perform on-par 3D editing with at least 20$\times$ faster speed per edit and be far easier to use in practice. It supports various types of editing, such as scene stylization, instance segmentation, and interactive drawing.
\bibliographystyle{splncs04}
\bibliography{ref.bib}

% \clearpage
\appendix
\renewcommand\thesection{\Alph{section}}
\renewcommand\thefigure{S\arabic{figure}}
\renewcommand\thetable{S\arabic{table}}
\renewcommand\theequation{S\arabic{equation}}
\setcounter{figure}{0}
\setcounter{table}{0}
\setcounter{equation}{0}

\begin{figure*}[t]
    % \vspace{-10pt}
    \begin{subfigure}{0.270\linewidth} % 213
        \includegraphics[width=1\linewidth]{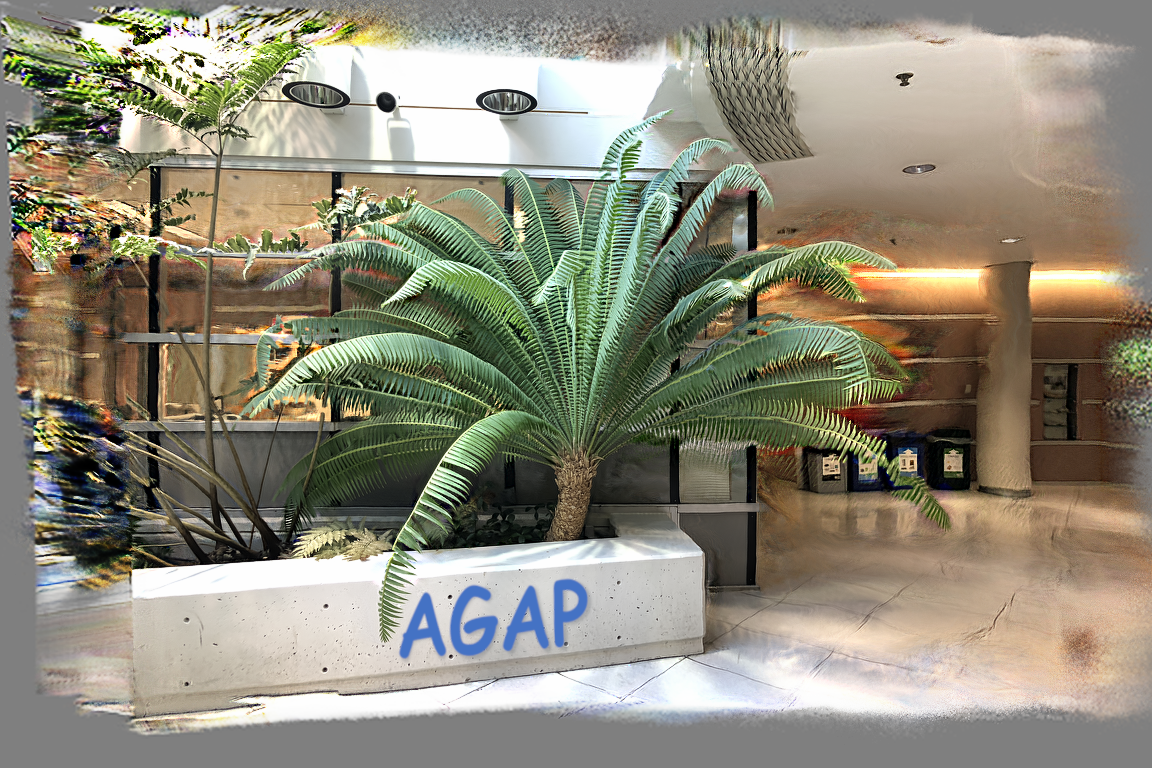}
    \end{subfigure}
    \begin{subfigure}{0.235\linewidth}
        \includegraphics[width=1\linewidth]{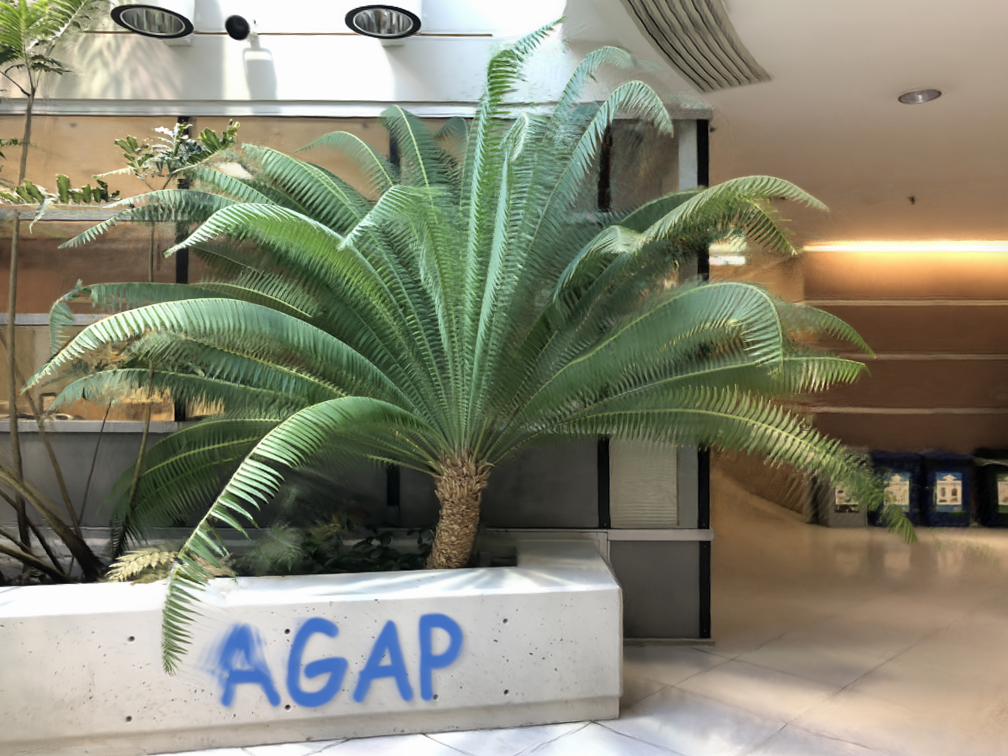}
    \end{subfigure}
    \begin{subfigure}{0.235\linewidth}
        \includegraphics[width=1\linewidth]{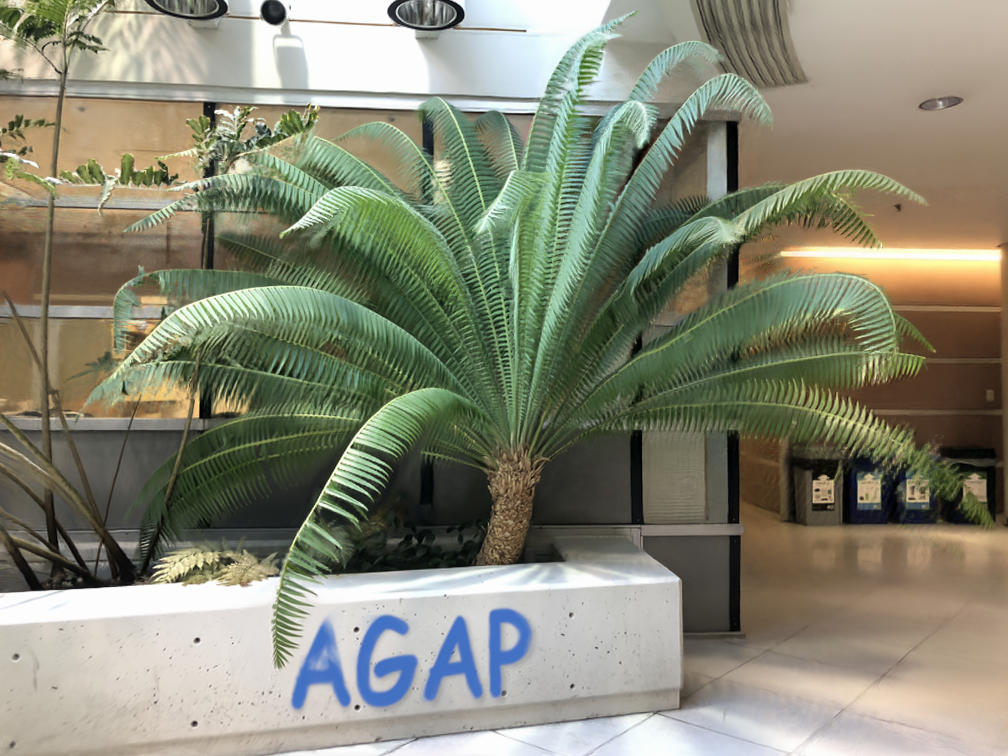}
    \end{subfigure}
    \begin{subfigure}{0.235\linewidth}
        \includegraphics[width=1\linewidth]{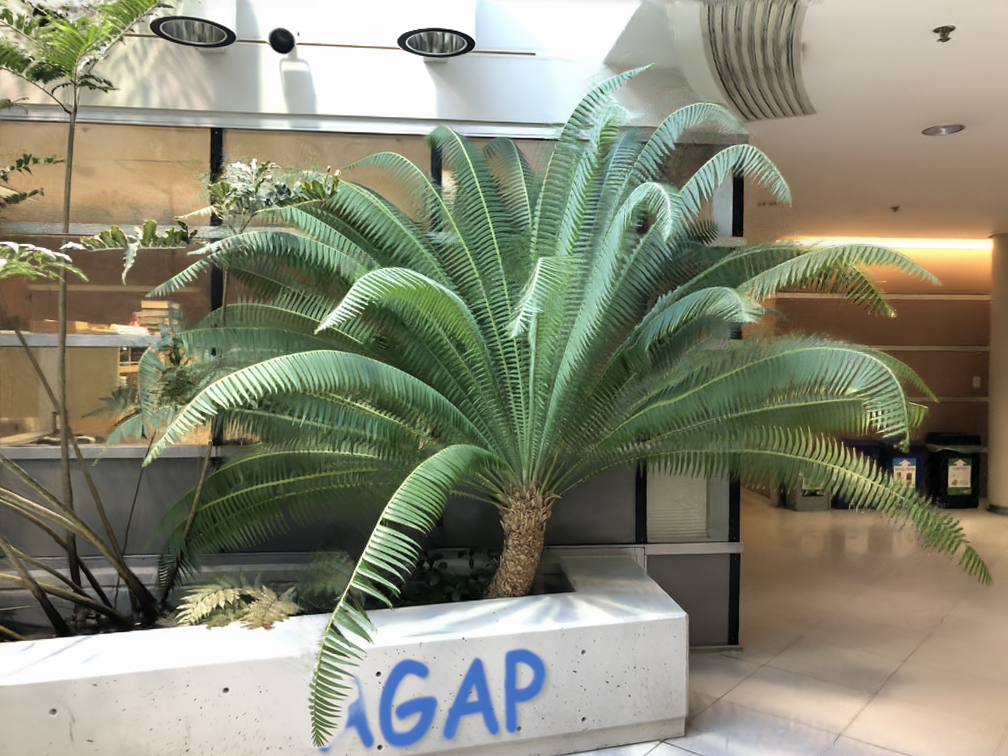}
    \end{subfigure}
    \begin{subfigure}{0.270\linewidth}\hfil Canonical with ``AGAP''\end{subfigure}
    \begin{subfigure}{0.235\linewidth}\hfil NVS 1 \end{subfigure}
    \begin{subfigure}{0.235\linewidth}\hfil NVS 2 \end{subfigure}
    \begin{subfigure}{0.235\linewidth}\hfil NVS 3 \end{subfigure}
    % \vspace{-18pt}
    \caption{Limitation of texture editing on occluded regions.}
    \label{fig:supp-limitation}
    % \vspace{-8pt}
\end{figure*}

\section{Supplementary Video}
To offer a more comprehensive demonstration of our visual results, we have included a supplementary video showcasing three editing cases (scene stylization, instance segmentation, and texture editing) on diverse 3D scenes. Please check ``demo.mp4'' for details.

\section{Failure Cases and Limitation}
Recall that in this paper, we come up with an editing-friendly representation, \method, which permits explicit 3D editing with the help of a natural 2D canonical image. In this section, we present some failure cases and discuss the limitation.

Our method supports texture editing by directly painting onto the canonical image. However, such painting might be distorted when the novel viewpoint exhibits occlusions on the edited regions. As shown in~\cref{fig:supp-limitation}, we can easily paint an ``AGAP'' logo onto the marble pedestal of the fern plant in the canonical image, allowing us to directly obtain the edited NVS from different novel viewpoints through neural rendering. However, the logo appears distorted in the regions that are occluded by the fern plant.

Our pipeline includes a projection offset $P_o$ to handle moderate levels of occlusion, which implicitly projects and clusters the 3D points to nearby pixels. However, we acknowledge that our method has limitations in handling extensive occlusions in the 3D scenes. Suppose a 3D scene is extremely complex and contains numerously extensive occlusions. Projecting such a scene onto a 2D plane (like a UV map) is possible, but creating a 2D projection that naturally and fully displays the scene for easy interactivity is very challenging and nearly impossible. Hence, such cases are beyond the scope of our current study, and we mainly focus on 3D scenes with moderate levels of occlusion.

\section{Training Details}
Our 3D editing pipeline involves a two-step process: (1) we first train a per-scene reconstruction model using the proposed \method representation, which includes an explicit density grid $\phi_G$, an explicit canonical image $\phi_I$, and an associated projection field $\phi_P$; (2) we can then perform explicit 2D edits on the canonical image $\phi_I$ for 3D scene editing, including scene stylization, instance segmentation, and texture editing. All experiments, including training on various scenes from different datasets, are conducted and tested on a single RTX A6000 GPU, with specific hyperparameter details outlined in~\cref{tab:supp-hyperparameters}.

\begin{table}[t]
    \centering
    \caption{
        \textbf{Hyperparameters} for training various scenes in different datasets.
    }
    \label{tab:supp-hyperparameters}
    % \vspace{-8pt}
    \SetTblrInner{rowsep=1.2pt}      % Row space.
    \SetTblrInner{colsep=4.0pt}      % Col space.
    \begin{tblr}{
        cells={halign=c,valign=m},   % Text alignment for all cells.
        column{1}={halign=l},        % Text alignment for the first column.
        cell{1}{1}={r=2}{},          % Multi row.
        cell{1}{2}={r=2}{},          % Multi row.
        cell{1}{3}={c=2}{},          % Multi column.
        hline{1,3,4,5,6,7,8}={1-4}{},     % Horizontal lines.
        hline{2}={3-4}{},     % Horizontal lines.
        hline{1,3,8}={1.0pt},          % Horizontal line width.
        vline{2,3}={1-7}{},   % Vertical lines.
    }
    Data Types                                                                                     & Image Size  & Weight Factor  &                \\ 
    \                                                                                              &             & $\lambda_{uv}$ & $\lambda_{tv}$ \\ 
    Forward-facing~\cite{mildenhall2019llff,haque2023instruct}                                     & (768, *)    & $10^{-5}$      & $10^{-5}$      \\ 
    Object-centric~\cite{mildenhall2020nerf,ma2022nep,jensen2014dtu}                               & (768, *) / (*, 768)  & $10^{-1}$      & $10^{-4}$      \\ Panorama~\cite{straub2019replica,habtegebrial2022somsi,haque2023instruct,barron2022mipnerf360} & (768, 1536) & $10^{-1}$      & $10^{-4}$      \\ 
    \end{tblr}
    % \vspace{-5pt}
\end{table}

\noindent\textbf{Optimization.} In the first stage, we employ the Adam optimizer~\cite{kingma2015adam} to optimize a per-scene model for $60$k steps with an initial learning rate of $0.1$ for both the explicit 3D density grid $\phi_G$ and 2D canonical image $\phi_I$, and a learning rate of $0.001$ for the implicit projection field $P$ with learnable parameter $\phi_P$. The optimization of the entire model involves an objective function comprising three main components: (1) an average $\mathcal{L}_2$ photometric loss $\mathcal{L}_{color}$ between the rendered pixel color $\hat{C}(\mathbf{r})$ and the ground-truth color $C(\mathbf{r})$; (2) a projection regularization $\mathcal{L}_{uv}$ aimed at minimizing the projection offset $\Delta\mathbf{p}_{uv}$; and (3) a total variation regularization applied to the density grid $\phi_G$.

\noindent\textbf{Weight factor.} As stated in the main paper, the final optimization process of our method to model the scene for efficient editing can be formulated as follows:
\begin{equation}
    \phi_G^*, \phi_I^*, \phi_P^* = \argmin_{\phi_G, \phi_I, \phi_P}{\mathcal{L}_{color} + \mathcal{L}_{uv} + \mathcal{L}_{tv}},
\end{equation}
where the second and the third terms are controlled by their corresponding weight factors $\lambda_{uv}$ and $\lambda_{tv}$, respectively. To be specific, the weight factor $\lambda_{uv}$ is set as $10^{-5}$ for forward-facing scene and larger value of $10^{-1}$ or $10^{-2}$ for panorama and inward-facing 360$^\circ$ scenes; the weight factor $\lambda_{tv}$ is set as $10^{-4}$ for panorama scene and $10^{-5}$ for other scenes. Note that for panorama and inward-facing 360$^\circ$ data, the total variation term is disabled after $20000$ steps to learn depths in detail.

\noindent\textbf{Progressive training.} Similar to~\cite{barron2022mipnerf360,mildenhall2020nerf,sun2022dvgo}, we apply progressive scaling for our voxel grid $\phi_G$ and canonical image $\phi_I$ for a coarse-to-fine learning process. By gradually refining the resolution of both representations, we enable a more detailed and comprehensive learning process.

At specific scaling-up milestone steps, we increase the $\phi_G$ voxel count by a factor of 2 and the $\phi_I$ pixel count by a factor of 4. For the forward-facing and object-centric data scenes, the voxel grid $\phi_G$ is scaled up at $\{2000, 4000, 6000, 8000\}$ training steps and the canonical image $\phi_I$ is scaled up at $\{8000, 16000\}$ training steps. For the panorama data types, the voxel grid $\phi_G$ is scaled up at $\{2000, 4000, 6000, 8000, 10000, 12000, 14000, 16000\}$ training steps, and the canonical image $\phi_I$ is scaled up at $\{4000, 8000, 12000, 16000\}$ training steps.

\noindent\textbf{Size of voxel grid $\phi_G$ and canonical image $\phi_I$.} After the progressive scaling up, The final resolution of the voxel grid $\phi_G$ is set as $384 \times 384 \times 256$ for forward-facing scenes and $320 \times 320 \times 320$ for other scenes.

For the NDC canonical camera of forward-facing scenes, we set the height $H_I$ of the learnable explicit canonical image $\phi_I$ as $768$, and the canonical image width $W_I$ is adaptively calculated according to the width-height aspect ratio of the training images and the computed bounding box of the scene in NDC space. Denoting the bounding box in NDC space as $(x'_{min}, x'_{max})$ in $x'$ dimension, $(y'_{min}, y'_{max})$ in $y'$ dimension, and $(z'_{min}, z'_{max}) = (-1, 1)$ in $z'$ dimension and the aspect ratio as $r_I$, we can then calculate the canonical image width as:
\begin{equation}
    W_I = H_I \times r_I \times \frac{x'_{max} - x'_{min}}{y'_{max} - y'_{min}}.
\end{equation}

For the canonical camera of panorama scenes, the canonical image height is set to be $768$ and the width $W_I$ is set to be $2 \times 768 = 1536$ according to the definition of Equirectangular projection. 

For the canonical camera of object-centric scenes, the canonical image width and height are adaptive, where the canonical image width-height aspect ratio $r_I = \tfrac{W_I}{H_I}$ is calculated according to the uv range:
\begin{equation}
    r_I = \frac{u_{max} - u_{min}}{v_{max} - v_{min}},
\end{equation}
where $u \in [-\pi, \pi]$ and $v \in [-\tfrac{\pi}{2}, \tfrac{\pi}{2}]$. The shorter dimension, whether width or height, is set to $768$.

\noindent\textbf{Annealed positional and hash encoding.} The projection offset employs Fourier positional encoding~\cite{vaswani2017attention} or multi-resolution hash encoding~\cite{muller2022instantngp} to capture high-frequency information. Given an input vector $\mathbf{x} \in \mathbb{R}^{3}$, the corresponding encoding can be defined as follows:
\begin{itemize}
    \item The positional encoding is defined as $\gamma_{pe} (\cdot): \mathbb{R}^3 \to \mathbb{R}^{3 \times (1 + 2K)}$ to encode 3-dimensional vector $\mathbf{x}$ up to $K$ frequencies as $\gamma_{pe}(\mathbf{x}) = [\mathbf{x}, F_1(\mathbf{x}), ..., F_K(\mathbf{x})]$. For the k-$th$ frequency of positional encoding, we have the encoding function $F_k(\mathbf{x}) = [\sin(2^k \mathbf{x}), \cos(2^k \mathbf{x})] \in \mathbb{R}^{2 \times 3}$. 
    \item The hash encoding is defined as $\gamma_{h} (\cdot): \mathbb{R}^3 \to \mathbb{R}^{3 + DK}$ to encode the vector $\mathbf{x}$ by a $K$-resolution hash grid with $D$-dimensional feature per layer as $\gamma_{h}(\mathbf{x}) = [\mathbf{x}, H_1(\mathbf{x}), ..., H_K(\mathbf{x})]$. For the $k$-th resolution hash grid with $D$-dimensional feature at each layer, we have the encoding function $H_k(\mathbf{x}) \in \mathbb{R}^{D}$.
\end{itemize}
Motivated by Nerfies~\cite{park2021nerfies}, the positional or hash encoding can incorporate an optional annealing learning strategy. Specifically, we introduce a weight factor $w^n_k = \frac{1}{2} (1 - \cos(\alpha^n_k \pi))$ for some encoded frequency $F_k^n$ or $H_k^n$ at some training step $n$, such that we have $F_k^n (\cdot) = w^n_k F_k(\cdot)$ or $H_k^n (\cdot) = w^n_k H_k(\cdot)$ and
\begin{equation}
    \alpha^n_k = \min(\max(\frac{n - N_{s}}{N_{e} - N_{s}} K - k, 0.0), 1.0),
\end{equation}
where $N_{s}$ and $N_{e}$ denote the start and end steps for anneal encoding, respectively. The strategy aims to facilitate the learning of low-frequency details and gradually incorporate high-frequency bands as the training progresses.

For all the experiments, the encoding $\gamma_{d}$ of direction $\mathbf{d}$ specifically employs positional encoding $\gamma_{pe}$, where we set $K = 4$ with the optional annealing learning strategy off. Concerning the encoding $\gamma_{p}$ of position $\mathbf{p}_{xyz}$, we choose to use positional encoding $\gamma_{pe}$ for PE models and hash encoding $\gamma_{h}$ for hash models, where we set $K = 8$ with the annealed learning starting at training step $N_s = 4000$ and ending at $N_e = 8000$ for PE models, and we set $D = 2$ and $K = 16$ without the optional annealed learning strategy for hash models.

\end{document}